\documentclass[reprint,onecolumn,amsmath,amssymb,jap,aip, nofootinbib]{revtex4-1}
\pdfoutput=1 
\usepackage{graphicx}
\usepackage{dcolumn}
\usepackage{bm}
\usepackage{epsfig}
\usepackage{mathptmx}
\usepackage{times}
\usepackage{amsmath}
\usepackage{amssymb}
\usepackage{dcolumn}
\usepackage{units}
\usepackage{upgreek}
\usepackage{tabularx}
\usepackage{threeparttable}
\usepackage[space]{grffile}
\usepackage[normalem]{ulem}

\usepackage[caption=false,font=footnotesize]{subfig}
\usepackage{float}
\usepackage{multirow}
\usepackage{array}
\usepackage{soul}
\usepackage{svg}
\usepackage{bbm}
\usepackage{verbatim}
\usepackage{multirow}
\usepackage{booktabs}
\usepackage{stmaryrd}
\usepackage{placeins}
\usepackage{hyperref}
\newtheorem{definition}{Definition}

\usepackage{xspace}

\newcommand\Included{\textcolor{teal}{\checkmark}}

\usepackage[acronym, style=super]{glossaries}

\newacronym{code}{GSBC}{generalized sparse block code}
\newacronym{llm}{LLM}{large language model}
\newacronym{rpm}{RPM}{Raven's progressive matrice}
\newacronym{sota}{SOTA}{state-of-the-art}

\newcommand{\longname}{Abductive Rule Learner with Context-awareness\xspace}
\newcommand{\name}{ARLC\xspace}
\newcommand{\fig}{Fig.}
\glsdisablehyper

\newcolumntype{C}{>{$}c<{$}}
\AtBeginDocument{
\heavyrulewidth=.08em
\lightrulewidth=.05em
\cmidrulewidth=.03em
\belowrulesep=.65ex
\belowbottomsep=0pt
\aboverulesep=.4ex
\abovetopsep=0pt
\cmidrulesep=\doublerulesep
\cmidrulekern=.5em
\defaultaddspace=.5em
}

\begin{document}
\title{Towards Learning to Reason: Comparing LLMs with Neuro-Symbolic on Arithmetic Relations in Abstract Reasoning}

\author{Michael Hersche} \email{michael.hersche@ibm.com} \affiliation{IBM Research -- Zurich}
\author{Giacomo Camposampiero} \affiliation{IBM Research -- Zurich}\affiliation{ETH Z\"{u}rich}
\author{Roger Wattenhofer} \affiliation{ETH Z\"{u}rich}
\author{Abu Sebastian}\affiliation{IBM Research -- Zurich}
\author{Abbas Rahimi}\affiliation{IBM Research -- Zurich}

\begin{abstract}

This work compares large language models (LLMs) and neuro-symbolic approaches in solving Raven's progressive matrices (RPM), a visual abstract reasoning test that involves the understanding of mathematical rules such as progression or arithmetic addition. 
Providing the visual attributes directly as textual prompts, which assumes an oracle visual perception module, allows us to measure the model's abstract reasoning capability in isolation. 
Despite providing such compositionally structured representations from the oracle visual perception and advanced prompting techniques, both GPT-4 and Llama-3 70B cannot achieve perfect accuracy on the \texttt{center} constellation of the I-RAVEN dataset. 
Our analysis reveals that the root cause lies in the LLM's weakness in understanding and executing arithmetic rules. 
As a potential remedy, we analyze the \longname (\name), a neuro-symbolic approach that \textit{learns to reason} with vector-symbolic architectures (VSAs).
Here, concepts are represented with distributed vectors s.t. dot products between encoded vectors define a similarity kernel, and simple element-wise operations on the vectors perform addition/subtraction on the encoded values.
We find that \name achieves almost perfect accuracy on the \texttt{center} constellation of I-RAVEN, demonstrating a high fidelity in arithmetic rules.
To stress the length generalization capabilities of the models, we extend the RPM tests to larger matrices (3$\times$10 instead of typical 3$\times$3) and larger dynamic ranges of the attribute values (from 10 up to 1000). 
We find that the LLM's accuracy of solving arithmetic rules drops to sub-10\%, especially as the dynamic range expands, while \name can maintain a high accuracy due to emulating symbolic computations on top of properly distributed representations. 
Our code is available at \url{https://github.com/IBM/raven-large-language-models}. 

\end{abstract}

\maketitle

\section{Introduction}
Abstract reasoning is often regarded as a core feature of human intelligence. 
This cognitive process involves abstracting rules from observed patterns in a source domain, and applying them in an unseen target domain. 
With the ultimate aim to achieve human-level intelligence, abstract reasoning tasks have sparked the interest of many in machine learning research.
Thanks to the availability of large datasets~\cite{barrett_measuring_2018, zhang_raven_2019, hu_stratified_2021}, various learning-based methods, ranging from pure connectionist~\cite{benny_scale-localized_2021, wu_scattering_2020} to neuro-symbolic~\cite{zhang_abstract_2021, hersche_neuro-vector-symbolic_2023, hersche_probabilistic_2024, camposampiero_towards_2024} approaches, achieved promising results in this domain.

More recently, the zero- and few-shot capabilities of \glspl*{llm} and their multi-modal variants have been tested on various abstract reasoning tasks such as verbal~\cite{webb_emergent_2023, stevenson_large_2023,gendron_large_2024-1, lewis_evaluating_2024} or visual~\cite{cao_what_2024,webb_emergent_2023, hu_-context_2023, mitchell_comparing_2024,camposampiero_abstract_2023,jiang_marvel_2024,ahrabian_curious_2024,zhang_how_2024, wust_bongard_2024, latif_systematic_2024_short,lewis_evaluating_2024} analogies.
One natural approach towards zero-shot visual abstract reasoning is to leverage multi-modal \gls{llm}'s vision capabilities to solve the task end-to-end. 
However, these multi-modal models perform significantly worse than their text-only version~\cite{mitchell_comparing_2024}, which might stem from a missing fine-grained compositional feature comprehension~\cite{cao_what_2024}. 
As an additional help, \glspl*{llm} have been provided with text-only inputs by giving them access to an oracle perception, i.e., providing perfectly disentangled representations~\cite{webb_emergent_2023, hu_-context_2023}. 
While this generally improves their reasoning abilities, \glspl*{llm} still fail to achieve perfect accuracy on many simple tasks. 
One example is represented by \glspl*{rpm}~\cite{raven_ravens_1938}, a benchmark that tests visual abstract reasoning capabilities by measuring the fluid intelligence of humans.
Here, the \gls*{sota} \gls*{llm}-based approach~\cite{hu_-context_2023} achieves only 86.4\% accuracy in the \texttt{center} constellation of I-RAVEN~\cite{hu_stratified_2021}, which we observe to be a gate-keeper for this task (see Section~\ref{sec:datasets}). 

In contrast, recent neuro-symbolic approaches showed not only almost perfect accuracy on the \texttt{center} constellation of I-RAVEN, but also demonstrated high fidelity in out-of-distribution (OOD) settings. 
For instance, the \longname (\name) represents attribute values with high-dimensional, distributed representations based on vector-symbolic architectures (VSAs)~\cite{plate_holographic_1995, plate_holographic_2003,gayler_vector_2003, kanerva_hyperdimensional_2009}. 
Learning the \gls*{rpm} rules boils down to a differentiable assignment problem of high-dimensional panel representations in a series of binding and unbinding operations, which can be solved with unconstrained optimization algorithms such as stochastic gradient descent (SGD).
\name outperformed the SOTA LLM-based approach~\cite{hu_-context_2023} both on in-distribution and OOD, thanks to relying on structured and similarity-preserving representations based on fractional power encoding (FPE)~\cite{plate_holographic_2003}. 

\begin{figure*}[t!]
    \centering
    \includegraphics[width=.88\textwidth]{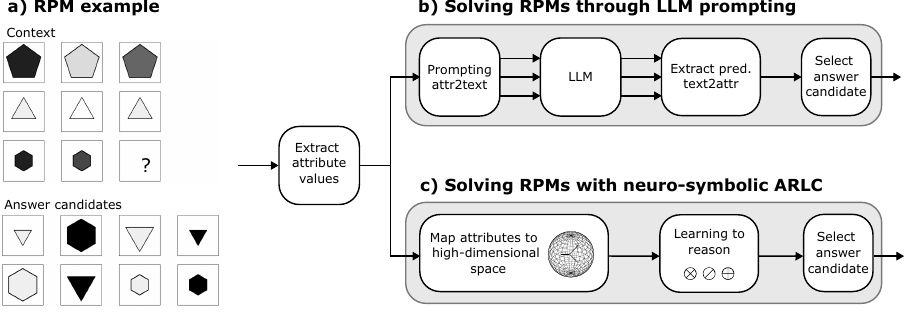}
    \caption{This work compares the abstract reasoning capabilities of \glspl*{llm} and neuro-symbolic \name on Raven's progressive matrices (RPM) tests. 
    \textbf{a)} An RPM example taken from the \texttt{center} constellation of I-RAVEN. The task is to find the empty panel at the bottom-right of the context matrix by selecting one of the answer candidates. 
    \textbf{b)} Solving RPMs through LLM prompting. Visual attribute values are extracted from the I-RAVEN dataset and assembled to individual per-attribute text-only prompts. \glspl*{llm} are prompted to predict the attribute of the empty panel. Finally, the attribute predictions are compared with the answer candidates, whereby the best-matching answer is selected as the final answer.
    \textbf{c)} Solving RPMs with neuro-symbolic \name that relies on distributed similarity-preserving representations and manipulates them via dimensionality-preserving operations; it learns rule-formulations as a differentiable assignment problem. 
    }
    \label{fig:main}
\end{figure*}

This paper extends on the initial work on \name~\cite{camposampiero_towards_2024}, by comparing its abstract reasoning capability with two prominent \glspl*{llm} (GPT-4~\cite{openai_gpt-4_2024_short} and Llama-3 70B~\cite{dubey_llama_2024_short}) (see \fig~\ref{fig:main}). 
Circumventing the perception by providing ground-truth attribute labels to the models allows us to measure their analogical and mathematical reasoning capabilities in isolation when such \emph{compositionally structured} (i.e., disentangled) representations are provided.
Our comprehensive prompting efforts lead to very high accuracy for Llama-3 70B (85.0\%) and GPT-4 (93.2\%), where the latter notably outperforms previous reports with GPT-3~\cite{hu_-context_2023} (86.4\%) and GPT-4 o1-preview~\cite{latif_systematic_2024_short} (18.00\%). 
This \gls*{llm}'s imperfect accuracy on the isolated task motivated us to further analyze their capability of detecting and executing different rules. 
In both GPT-4 and Llama-3 70B, we find a notable weakness in performing arithmetic rules that require row-wise additions or subtractions (e.g., see the last prompt in \fig~\ref{fig:extended_raven}).   
To gain more insight about this behavior, we set up a new RPM dataset (I-RAVEN-X) that increases the grid size from 3$\times$3 to 3$\times$10, additionally allowing for a configurable dynamic range for the arithmetic computations. 
Also here, we observe a notable weakness in the arithmetic rule that gets even amplified by an increasing dynamic range.
On the other hand, \name demonstrates high accuracy on larger grid sizes and allows to increase the dynamic range without further retraining, thanks to the the capability of adjusting the underlying structured FPE representations.


\section{Datasets}

\subsection{I-RAVEN}\label{sec:datasets}
We test the models on the \texttt{center} constellation of I-RAVEN~\cite{hu_stratified_2021} (see \fig~\ref{fig:main}). 
The test consists of a 3$\times$3 context matrix and eight answer candidate panels. 
Each panel contains an object, characterized by different attributes (shape, size, and color). 
The relation between each attribute's value in different panels is governed by a well-defined set of rules: \texttt{constant}, \texttt{progression}, \texttt{arithmetic}, and \texttt{distribute three}. 
The task is to infer the rule governing each attribute in the context matrix and use it to determine the content of the missing (bottom-right) panel, selecting it within the eight candidate answers. 
Compared to other \gls*{rpm} benchmarks that have been used to evaluate \glspl*{llm}~\cite{webb_emergent_2023}, I-RAVEN tests a more complex range of logical and arithmetic skills.
While I-RAVEN provides tests in various constellations with more objects that may intuitively appear more arduous to solve, \glspl*{llm} are more challenged with the seemingly simple constellations. 
For instance, GPT-3 achieved a higher accuracy on the \texttt{2x2} and \texttt{3x3} constellations (78.0\% and 86.4\%) than on \texttt{center} (77.2\%)~\cite{hu_-context_2023}.
Moreover, high accuracy can be maintained on the \texttt{2x2} and \texttt{3x3} constellations while only looking at the last row of the context matrix~\cite{hu_-context_2023}, effectively showing that no analogical reasoning is required to solve the test in these constellations. 
Hence, we opted to focus our evaluation on the \texttt{center} constellation only, using 500 samples from I-RAVEN's test set.  
Inspired by recent works~\cite{webb_emergent_2023, hu_-context_2023}, we simplify \gls*{rpm} from a visual abstract reasoning test to a purely abstract reasoning test. 
Assuming a perfect perception, we extract the attribute values from I-RAVEN and use them to create the prompts for the model.

\subsection{New I-RAVEN-X}
To further evaluate the mathematical reasoning capabilities at scale, we introduce an extension of the I-RAVEN's \texttt{center} constellation, called I-RAVEN-X. 
Our new benchmark maintains I-RAVEN's rules and attributes but allows for a parameterizable number of columns ($g$) and a dynamic range of attribute values ($m$). 
When generating a new \gls*{rpm} example, we uniformly sample from one of the available rules (\texttt{constant}, \texttt{progression}, \texttt{arithmetic}, and \texttt{distribute three}). 
Note that the attribute \texttt{shape} does not incur the \texttt{arithmetic} rule. 
We use I-RAVEN's attribute bisection tree~\cite{hu_stratified_2021} to generate unbiased candidate answers. 

\begin{figure*}
    \centering
    \includegraphics[width=.9\linewidth]{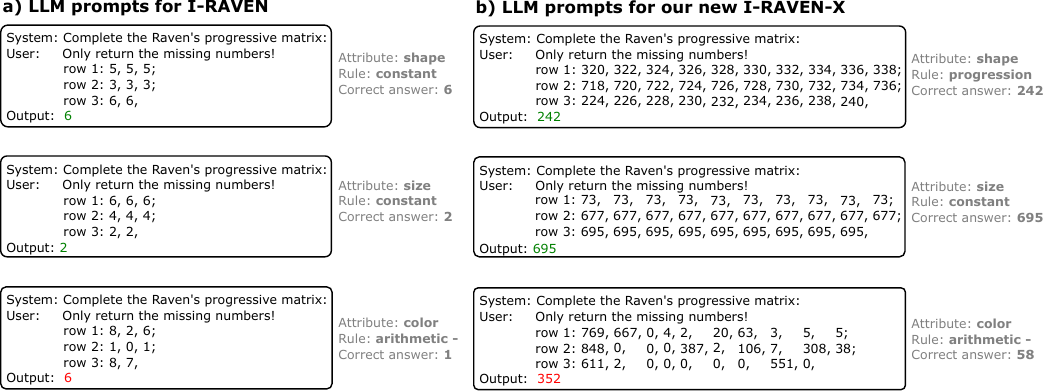}
    \caption{\textbf{a)} Individual per-attribute text-only prompts to solve \gls*{rpm} tasks from I-RAVEN. \textbf{b)} Example prompts with of our novel configurable I-RAVEN-X dataset of size 3$\times$10 with a value range of $m=1000$. In both the I-RAVEN and I-RAVEN-X examples, the \gls*{llm} (GPT-4) errs in the arithmetic rules.}
    \label{fig:extended_raven}
\end{figure*}

In the following, we describe the context matrix generation for the individual rules. 
The overall goal is that the values stay in the range $[ 0, m-1]$. 
\begin{itemize}
\item \texttt{constant}: This rule keeps the attribute value constant per row. For each row, we uniformly sample an integer from the set $\lbrace 0, 1, ..., m-1 \rbrace$, and duplicate along the row. 
\sloppy
\item \texttt{progression}: The attribute value monotonically increases or decreases in a row by a value of 1 or 2. First, we uniformly sample the progressive increment/decrement ($\delta$) from the set $\lbrace -2, -1, +1, +2 \rbrace$. In case of a positive increment, we first define the values of the right-most columns, by uniformly sampling from the set $\lbrace (g-1)\cdot \delta, ..., m-1  \rbrace$ for each row.
Then, the rest of the matrix is completed by applying the progression rule. 
The sampling for a negative $\delta$ is done specularly from the first column. 
\item \texttt{arithmetic}: The attribute values of the first $g-1$ panels are either added (\texttt{arithmetic plus}) or subtracted (\texttt{arithmetic minus}), yielding the attribute value of the last panel in the row. 
In \texttt{arithmetic plus}, we sequentially sample the values from the first $g-1$ panels in the row. For each panel, we set the sampling range to $\lbrace 0, ..., m - s \rbrace$, where $s$ is the sum of the already sampled panels in the row. 
Afterward, the first $g-1$ panels are shuffled. 
Finally, the values of the last panels are the sum of the first $g-1$ ones, applied row-wise. 
For \texttt{arithmetic minus}, we apply the same sampling strategy but leave the first column empty. 
The value of the first column is then defined as the sum of the other columns.
\item \texttt{distribute-n}: We uniformly sample distinct values for the first row from $\lbrace 0, ..., m-1  \rbrace$. 
The content of the remaining rows is defined by applying a circular shift per row (either right or left). 
\end{itemize}
\fig~\ref{fig:extended_raven}b shows example prompts generated from samples of our new dataset. 

\section{LLM-based RPM solving}

\subsection{Models}
We focused our evaluations on text-only \glspl*{llm}. 
There exist attempts~\cite{mitchell_comparing_2024,jiang_marvel_2024, cao_what_2024, ahrabian_curious_2024, zhang_how_2024} that leverage vision support of some multi-modal \glspl*{llm} (e.g., GPT-4V) directly feeding the models with visual \gls*{rpm} data; however, they achieve consistently lower reasoning performance than with text-only prompting. 
The \gls*{sota} \gls*{llm}-based abstract reasoning approach~\cite{hu_-context_2023} relied on reading out GPT-3's (\texttt{text-davinci-002}) token probabilities.
However, this model is no longer accessible to users and its successive iterations do not allow the retrieval of prediction logits.
Hence, we considered discrete classification approaches that are based on output strings rather than distribution over tokens.
In particular, we investigated two \gls*{sota} \glspl*{llm}: the proprietary GPT-4~\cite{openai_gpt-4_2024_short}(\texttt{gpt-4-0613})\footnote{GPT-4 was accessed between 07/03/2024--10/30/2024. 
}  and the open-source Llama-3\footnote{The model weights were downloaded and evaluated locally.} 70B~\cite{dubey_llama_2024_short}.
More recent iterations of these models were not considered in our analysis for different reasons.
Meta's attribution requirement in their updated terms regarding naming conventions prevented us from testing Llama-3.1
During initial tests, GPT-4o yielded worse results than GPT-4, hence we focused on GPT-4. 
Moreover, GPT-4 o1's poor abstract reasoning results on \gls*{rpm}~\cite{latif_systematic_2024_short} (18\% on \texttt{2x2} RAVEN) and its limited availability (only preview version available) prevented us from performing statistically significant tests on this chain-of-thought model. 

\subsection{Prompting and classification}

\paragraph{Entangled and disentangled prompts}
Following~\cite{hu_-context_2023}, we evaluated two different prompting strategies, \textit{entangled} and \textit{disentangled} prompting.
The entangled prompting provides all the attributes' values in a single prompt (see Appendix~\ref{app:disentanglement}).
The disentangled prompting, on the other hand, is a compositionally structured approach that queries the \gls*{llm} for individual attribute prediction. 
Disentangled prompting simplifies the task, but increases the number of queries by 3$\times$. 

\paragraph{Discriminative and predictive classification}
Similarly to~\cite{gendron_large_2024-1}, we consider two approaches to solve RPM tests with \glspl*{llm}. 
In the \textit{discriminative} approach, we provide the attribute descriptions of both the context matrix and the answer candidates. 
The \gls*{llm} is then asked to return the panel number of the predicted answer. 
Appendix~\ref{app:discriminative} provides an example prompt of the discriminative approach. 
In the \textit{predictive} approach, we prompt the \gls*{llm} only with the context matrix without the candidate answers. 
The \gls*{llm} has to {predict} the value of the empty panel (see \fig~\ref{fig:extended_raven}). 
For selecting the final answer, we compare the predicted values with the answer panels and pick the one with the highest number of overlapping values. 
While the predictive approach may appear more difficult, it implicitly biases the \gls*{llm} to approach the task as humans usually do, i.e., first applying a generative process to abduce rules and execute them to synthesize a possible solution, and then discriminatively selecting the most similar answer from choices~\cite{holyoak_oxford_2013}.
Moreover, the final answer selection is done without the intervention of the \gls*{llm}, rendering phenomena like hallucinations less likely.
Thus, the predictive classification can be seen as a more guided approach that helps \gls*{llm} to solve the task. 

\paragraph{Self-consistency}
As an optional extension, we employ self-consistency~\cite{wang_self-consistency_2023, lewkowycz_solving_2022} by querying the model multiple times ($n=7$ times), sampling the next token from the distribution with a non-zero soft-max temperature.
We find the optimal soft-max temperature for GPT-4 ($T=0.5$) and Llama-3 70 B ($T=0.4$) via a grid search on a subset of 50 I-RAVEN problems. 
We did not explore the effect of other parameters, such as top-k or top-p, and set them to the default values. 
The final prediction is determined by a majority vote over the sampled outputs.  
The selection of an odd number of samples (i.e., $n=7$) helps to prevent potential ties. 

\paragraph{In-context learning}
For a better understanding of the RPM task, we optionally prefix 16 in-context examples to the prompt~\cite{brown_language_2020}. 
In the predictive classification approach (where no answer candidates are provided), we simply provide complete example RPM matrices. 
The in-context samples are randomly selected from I-RAVEN's training set. 
Examples that had the same context matrix as the actual task are discarded and re-sampled to prevent shortcut solutions.

\section{\name: learning abductive reasoning using VSA distributed representations}

\begin{figure*}
    \centering
    \includegraphics[width=.9\linewidth]{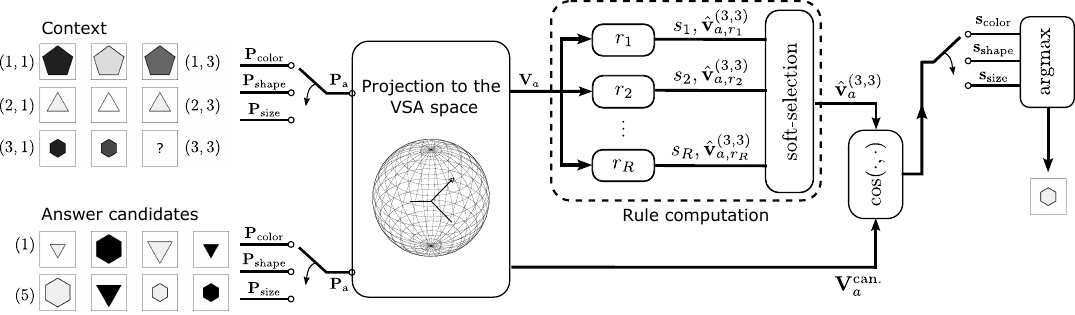}
    \caption{\name architecture. \name maps attribute values, or distributions of values, to distributed VSA representations, where the semantic similarity between values is preserved via a notion of kernel. Learnable rules ($r_1, ..., r_R$) predict the VSA representation of the empty panel ($\hat{\mathbf{v}}_{a, r}^{(3,3)}$) together with a confidence value ($s_r$). 
    The closest answer to the predicted soft-selected prediction ($\hat{\mathbf{v}}_a^{(3,3)}$) is chosen as the final answer. 
    }
    \label{fig:arlc}
\end{figure*}

This section presents the \longname (\name), which performs neuro-symbolic reasoning with distributed VSA representations (see \fig~\ref{fig:arlc}). 
\name projects each panel's attribute value (or distributions of values) into a high-dimensional VSA space. 
The resulting VSA vectors preserve the semantic similarity between attribute values: the dot products between corresponding VSA encoded vectors define a similarity kernel~\cite{plate_holographic_2003, frady_computing_2022}. 
Moreover, simple component-wise operations on these vectors, binding and unbinding, perform addition and subtraction respectively on the encoded values.
For rule learning, \name introduces a generic rule template with several terms forming a series of binding and unbinding operations between vectors.
The problem of learning the rules from data is reduced to a differentiable assignment problem between the terms of the general rule template and the VSA vectors encoding the contents of the panels, which can be learned with standard SGD. 
\name was initially presented in~\cite{camposampiero_towards_2024}; this work mainly compares it to the reasoning capabilities of LLMs on I-RAVEN, and demonstrates its extension to larger grid sizes and dynamic ranges on our novel I-RAVEN-X. 

\begin{figure}
    \centering
    \includegraphics[width=0.4\linewidth]{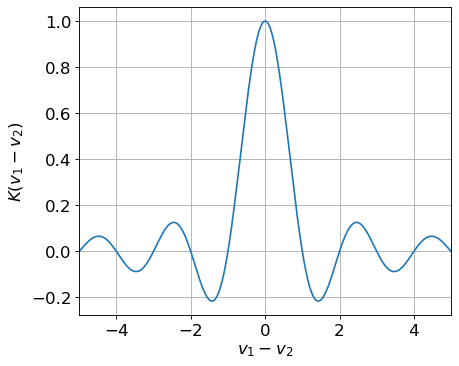}
    \caption{Similarity kernel in VSA. Mapping two values ($v_1$ and $v_2$) to a VSA space (i.e., GSBC in \name) that uses fractional power encoding (FPE) and computing their similarity in the VSA space yields the shown similarity kernel $K(v_1-v_2)$.}
    \label{fig:sbc-kernel}
\end{figure}

\subsection{From visual attributes to distributed VSA representations}

\name's key concept is to represent attribute values with high-dimensional, distributed VSA vectors that preserve the semantic similarity between the attribute values thanks to an introduced kernel notion. 
We start by defining a VSA that equips the space with dimensionality-preserving vector operations (binding $\otimes$, unbinding $\oslash$, and bundling $\oplus$) as well as a similarity function ($\mathrm{cos}(\cdot, \cdot)$). 
For example, \name uses binary generalized sparse block codes (GSBCs)~\cite{hersche_factorizers_2024} as a particular VSA instance. In binary GSBCs, the $D$-dimensional vectors are divided into $B$ blocks of equal length, $L=D/B$, where only one (randomly selected) element per block is set to 1 ($D=1024$ and $B=4$).
The algebraic operations of binary GSBCs are defined in Table~\ref{tab:vecop}.
See Appendix~\ref{app:vsa} for a detailed background on VSA. 

Next, we define a mapping $z: \mathbb{Z}^+ \rightarrow \mathbb{R}^D$ that enables the projection of input RPM attributes into a corresponding high-dimensional, semantically-rich feature space.
Note that this work focuses on mapping integer values as the attribute values in I-RAVEN are integer-valued too. However, generalizing this approach to real-valued domain mappings is possible using frequency holographic reduced representations (FHRR)~\cite{plate_holographic_1995}.
Leveraging fractional power encoding (FPE)~\cite{plate_holographic_2003}, a value $v\in \mathbb{Z}^+$ is encoded as
\begin{align*}
\mathbf{z}(v) = \mathbf{z}^v = \bigotimes_{n=1}^v \mathbf{z},
\end{align*}
where $\mathbf{z} \in \mathbb{R}^D$ is a randomly drawn binary GSBC vector. 
This mapping yields a similarity kernel between neighboring vector representations~\cite{frady_computing_2022}, as shown in \fig~\ref{fig:sbc-kernel}.

Let us assume two variables with values $v_1$ and $v_2$, which are represented with two VSA vectors ($\mathbf{z}(v_1)= \mathbf{z}^{v_1}$ and $\mathbf{z}(v_1)= \mathbf{z}^{v_2}$). 
Binding the two vectors yields $\mathbf{z}(v_1) \otimes \mathbf{z}(v_2)  = \mathbf{z}^{v_1} \otimes \mathbf{z}^{v_2} = \mathbf{z}^{v_1+v_2}$. 
Hence, binding in the VSA space is equivalent to the addition in $\mathbb{R}$. 
In other words, the FPE initialization allows to establish a semantic equivalence between high-dimensional vectors and real numbers.
This property is consistently exploited in \name's framework, as it allows to solve the analogies in the \gls*{rpm} puzzles as simple algebraic operations in the domain of real numbers.
For example, by computing the similarity between the bound representation and a third projected variable ($\mathrm{sim}(\mathbf{z}^{v_1+v_2},\mathbf{z}^{v_3}) $), we can evaluate if $v_1+v_2 \stackrel{?}{=} v_3$ representing the \texttt{arithmetic plus} rule in \gls*{rpm}. 

One advantage of performing reasoning with distributed VSA representations is its capability to represent perceptual uncertainty in the variable values. 
Connecting to the previous example, let us assume that the first variable takes value $v_1$ with probability $p$ and value $v_1'$ with probability $p'=1-p$.
The distribution can be encoded as the weighted superposition of the two corresponding codewords: $p \cdot \mathbf{z}^{v_1} + p' \cdot \mathbf{z}^{v_1'}$. 
The similarity computation between the bound representation and a third variable would then yield 
\begin{align*}
    \mathrm{sim}((p\cdot \mathbf{z}^{v_1} + p' \cdot \mathbf{z}^{v_1'})\otimes \mathbf{z}^{v_2} ,\mathbf{z}^{v_3}) &=  \mathrm{sim}(p\cdot \mathbf{z}^{v_1} \otimes \mathbf{z}^{v_2} + p' \cdot \mathbf{z}^{v_1'}\otimes \mathbf{z}^{v_2}  ,\mathbf{z}^{v_3}) \\
    &= p \cdot \mathrm{sim}(\mathbf{z}^{v_1} \otimes \mathbf{z}^{v_2} ,\mathbf{z}^{v_3}) + p' \cdot \mathrm{sim}(\mathbf{z}^{v_1'} \otimes \mathbf{z}^{v_2} ,\mathbf{z}^{v_3}), 
    \end{align*}
where the first equality uses the linearity of the binding operation, and the second one the linearity of the similarity metric. 
Overall, this formulation allows the validation of multiple solutions (in this case two) using only a single binding and similarity computation.

In the RPM application, each panel's label is translated to a probability mass function (PMF) $\mathbf{p}_a^{(i,j)}$, where $a$ is the attribute, $i$ is the row index and $j$ is the column index of the panel.
The panel's PMF is then projected into the VSA space as
\begin{align*}
    \mathbf{v}_a^{(i,j)} = \sum_{k=1}^m \mathbf{p}_a^{(i,j)}\left[k\right] \cdot \mathbf{z}^k, 
\end{align*}
where $m$ is the number of possible values that the attribute $a$ can assume.
Overall, this yields eight VSA vectors for each attribute $a$ (one for each panel of the input RPM matrix), represented by
\begin{align}
 \mathbf{V}_a := \left( \mathbf{v}_a^{(1,1)}, \mathbf{v}_a^{(1,2)}, \dots, \mathbf{v}_a^{(3,2)}\right). 
\end{align}
Note that the basis vectors are pre-computed and stored in a dictionary $\mathbb{C}=\{\mathbf{z}^k \}_{i=1}^{r}$ containing $m$ elements. 

    \begin{table}[t]
    \caption{Supported VSA operations and their equivalent in $\mathbb{R}$.} 
    \label{tab:vecop}
    \centering
    \begin{tabular}{ccc} 
    \toprule
    Operation & Binary GSBCs with FPE & Equivalent in $\mathbb{R}$ \\
    \cmidrule(r){1-1}\cmidrule(r){2-2}\cmidrule(r){3-3}
    Binding ($\otimes$) & Block-wise circular convolution & Addition $+$\\
    Unbinding ($\oslash$) & Block-wise circular correlation & Subtraction $-$\\
    Bundling ($\oplus$) & Sum \& normalization & ---\\
    Similarity ($\odot$) & Cosine similarity ($\text{cos}(\cdot,\cdot)$) & ---\\
    \bottomrule
    \end{tabular}
    \end{table}

\subsection{Learning RPM rules as an assignment problem}
As we have seen in the previous example, \gls*{rpm} rules can be represented using VSA operations. 
Generalizing the application beyond the \texttt{arithmetic plus} rule, we find that the rules used in RPM can be framed as a series of binding and unbinding operations:
    \begin{align}
        \label{eq:newtemp}
        r = \left( \mathbf{c}_1 \otimes \mathbf{c}_2 \otimes \mathbf{c}_3 \otimes \mathbf{c}_4 \otimes \mathbf{c}_5 \otimes \mathbf{c}_6 \right) \oslash \left( \mathbf{c}_7 \otimes \mathbf{c}_8 \otimes \mathbf{c}_9 \otimes \mathbf{c}_{10} \otimes \mathbf{c}_{11} \otimes \mathbf{c}_{12} \right),
    \end{align}    
where $\mathbf{c}_i$ represents a context panel $\mathbf{v}_a^{(i,j)}$ or the identity $\mathbf{e}$. 
In this setting, learning the rules of RPM can hence be interpreted as an assignment problem between VSA vectors and terms of Equation~\eqref{eq:newtemp}.

\begin{figure*}[t]
\centering
\subfloat[$X=\{R_1\}, O=\{R_2,R_3\}$]{\includegraphics[width=0.23\textwidth]{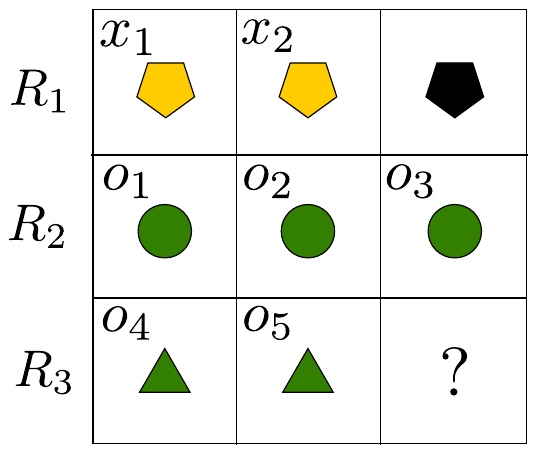}\label{fig:context1}}
\qquad
\subfloat[$X=\{R_2\}, O=\{R_1,R_3\}$]{\includegraphics[width=0.23\textwidth]{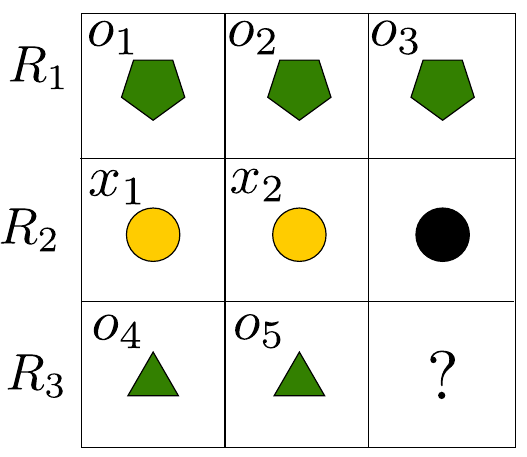}  \label{fig:context2}}  
\qquad
\subfloat[$X=\{R_3\}, O=\{R_1,R_2\}$]{\includegraphics[width=0.23\textwidth]{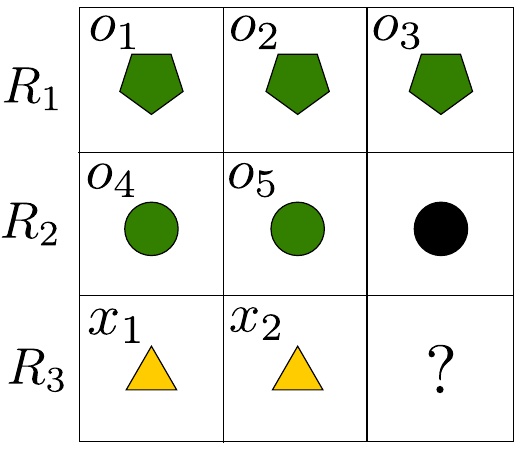}\label{fig:context3}}
\caption{Visualization of current samples ($X=\{x_1,x_2\}$, in yellow) and context ($O=\{o_1,\dots,o_5\}$, in green) panels when predicting the third panel for different rows, namely the first row (left), second row (center) and third row (right). Black objects represent panels that are not used for the computation, while the question mark represents the unknown test panel, which is unavailable during inference.}
\label{fig:context}
\end{figure*}

Motivated by works in cognitive sciences and psychology that argue for the importance of context in the solution of analogies for humans ~\cite{chalmers_high-level_1992,cheng_context-dependent_1990}, \name uses a general formulation of the soft-assignment problem which relies on the notion of \textit{context}: 
\begin{align}
    \label{eq:newopti}
   \mathbf{c}_k = \sum_{i=1}^I w_k^i \cdot \mathbf{x}_i  + \sum_{j=1}^J u_k^j \cdot \mathbf{o}_j  +  v_k\cdot \mathbf{e},
\end{align}
where $\mathbf{w},\mathbf{u},\mathbf{v}$ are the learned parameters and they are subject to the following constraints:

\begin{small}
    \begin{align*}
       \sum_{i=0}^I w_k^i + \sum_{j=0}^J w_k^j +  v_k = 1, \quad 0 \leq w_k^i \leq 1 \, \forall i, \quad 0 \leq u_k^j \leq 1 \, \forall j, \quad 0 \leq v_{k} \leq 1, \, \forall k.
    \end{align*}
\end{small}
Here, $\mathbf{X}=\{\mathbf{x}_1, \dots, \mathbf{x}_I\}$ is the set of attributes that define the current sample, that is, the description of the problem for which we infer a solution.
$\mathbf{O}=\{\mathbf{o}_1, \dots, \mathbf{o}_J\}$ is the set of attributes that define the context for that sample, that could be interpreted as a working memory from which additional information to infer the answer can be retrieved. 
For predicting the empty panel in the last row, the context ($\mathbf{O}$) corresponds to the first two rows and the current samples ($\mathbf{X}$) to the last row (see \fig~\ref{fig:context3}). 
We augment this standard prediction with two more permutations, which aim to predict the last panel of the first and second row (see \fig~\ref{fig:context1} and \fig~\ref{fig:context2}). 
The knowledge of the right-most panels in the first two rows allows us to compute a rule confidence by comparing the rule's prediction with the actual panel representation via the cosine similarity.

\subsection{Executing and selecting the learned rules}
Inference with the learned rule set is a two-step process: an execution step (where all the rules are applied in parallel to the input) and a selection step (where a prediction for the missing panel is generated).
The application of each rule $r$ to an RPM example generates a tuple of three VSA vectors $(\hat{\mathbf{v}}_{a,r}^{(i,3)})^3_{i=1}$, which corresponds to the result of the rule execution on the three rows of the RPM matrix, together with a rule confidence value $s_r$.
The confidence value is computed as the sum of the cosine similarities between the predicted VSA vectors and  their respective ground-truth vector,
\begin{align}
\label{eq:conf}
    {s}_r =  \sum_{i=1}^3 \mathrm{cos}\left(\mathbf{v}_a^{(i,3)},  \hat{\mathbf{v}}_{a,r}^{(i,3)} \right).
\end{align}
During inference, the last term of the sum ($i=3$) is omitted, as the ground truth for the third row is unknown.

The answer is finally produced by taking a linear combination of the VSA vectors generated by executing all the rules, weighted by their respective confidence scores (normalized to a valid probability distribution using a softmax function).
More formally, if we define $\mathbf{s}=\left[ s_1, \dots, s_R \right]$ to be the concatenation of all rules' confidence score and $\hat{\mathbf{V}}_a^{(3,3)} = [ \hat{\mathbf{v}}_{a,1}^{(3,3)}, \dots,  \hat{\mathbf{v}}_{a,R}^{(3,3)} ]$ to be the concatenation of all rules' predictions for the missing panel, the final VSA vector predicted by the model for the attribute $a$ becomes
\begin{align}
\label{eq:softsel}
    \hat{\mathbf{v}}_a^{(3,3)} = \text{softmax}\left(\mathbf{s}\right) \cdot \hat{\mathbf{V}}_a^{(3,3)}. 
\end{align}
The use of the weighted combination can be understood as a \textit{soft-selection} mechanism between rules and was found to be more effective compared to the \textit{hard-selection} mechanism provided by sampling~\cite{hersche_probabilistic_2024}. 

\subsection{Training Loss and other Implementation Aspects}
We follow the training recipe provided by Learn-VRF~\cite{hersche_probabilistic_2024}.
The model is trained using stochastic gradient descent (SGD) with a learning rate $\text{lr}=0.01$ for 25 epochs.
The training loss is defined as the inverse cosine similarity between the three predicted  panels  and their corresponding ground truth
\begin{align}
    \mathcal{L}= 1- \sum_{i=1}^3 \mathrm{cos}\left(\mathbf{v}_a^{(i,3)},  \hat{\mathbf{v}}_a^{(i,3)}\right).
\end{align}

As in Learn-VRF, we set the number of rules to $R=5$.
A single set of rules is instantiated and shared between all RPM attributes.

\subsection{Applying \name on I-RAVEN-X}
While \name was initially designed for I-RAVEN, it can be seamlessly extended to our I-RAVEN-X with minor modifications. 
First, the number of binding/unbinding terms in Equation~\eqref{eq:newtemp} is increased, e.g., from 12 to 22 to support the larger grid size of $g=10$. 
Moreover, we increase the number of entries in the dictionary ($\mathbb{C}$) to support the larger dynamic range ($m$). 
Notably, only varying the dynamic range at constant grid size does not require retraining: we can simply replace the dictionary in order to support OOD generalization. 
Indeed, we could demonstrate that \name trained on a dynamic range of $m=45$ can favorably generalize to a dynamic range of $m=1000$. 

\section{Results}

\subsection{Main results on I-RAVEN}

    \begin{table}[t]
    \centering
    \caption{Task accuracy (\%) on the \texttt{center} constellation of I-RAVEN. Among the baselines, we replicate Learn-VRF~\cite{hersche_probabilistic_2024}; the other results are taken from~\cite{hersche_neuro-vector-symbolic_2023}. The standard deviations are reported over 10 random seeds. Llama-3 and GPT-4 are queried with the corresponding best prompting technique (see Table~\ref{tab:llm-ablation}). Number of parameters for GPT-4 is not publicly available. The reasoning backend of PrAE, NVSA, and our $\text{\name}_{\text{p}\mapsto\text{l}}$ do not have trainable parameters.} 
    \label{tab:idresults}
    \begin{tabular}{ l c l} 
    \toprule
    Method & Parameters &  Accuracy \\
    \cmidrule(r){1-2}\cmidrule(r){3-3}
    MLP~\cite{hersche_probabilistic_2024}& 300\,k& 97.6 \\
    SCL~\cite{wu_scattering_2020} & 961\,k  &  ${99.9^{\pm 0.0 }}$ \\
    PrAE~\cite{zhang_abstract_2021} & n.a. &  $83.8^{\pm 3.4 }$ \\
    NVSA~\cite{hersche_neuro-vector-symbolic_2023}& n.a.& $99.8^{\pm 0.2 }$ \\
    Learn-VRF~\cite{hersche_probabilistic_2024}& 20\,k& $97.7^{\pm4.1}$  \\
    GPT-3~\cite{hu_-context_2023}& 175\,b  & 86.4  \\
    \cmidrule(r){1-2}\cmidrule(r){3-3}
    Llama-3 & 70\,b & 85.0\\
    GPT-4 & unk. & 93.2\\
    $\text{\name}_\text{progr}$& n.a.&  $97.2^{\pm 0.0}$ \\
    $\text{\name}_{\text{p}\mapsto\text{l}}$ & 480&  $97.6^{\pm 0.0}$  \\
    $\text{\name}_\text{learn}$& 480&  $98.4^{\pm 1.5}$  \\
    \bottomrule
    \end{tabular}
    \end{table}

Table~\ref{tab:idresults} compares our LLM results with \name on the \texttt{center} constellation of I-RAVEN, considering also a range of neuro-symbolic and connectionist baselines.
For the LLMs, we show the results with the corresponding best prompting techniques (see the ablation in Section~\ref{sec:llm-ablation}). 
Moreover, we present results for three different versions of \name: $\text{\name}_\text{progr}$, where the model's weights are manually programmed with RPM rules ($R=4$, since \texttt{constant} can be considered as a special case of \texttt{progression}), $\text{\name}_{\text{p}\mapsto\text{l}}$, where the model is initialized with the programmed rules and then trained with gradient descent, and $\text{\name}_\text{learn}$, where the rules are learned from scratch from data.

Among the LLM approaches, our GPT-4-based approach achieved the highest accuracy (93.2\%) notably outperforming previous \gls*{sota} \gls*{llm}-based abstract reasoning approaches on this benchmark (86.4\%)~\cite{hu_-context_2023}. 
Yet, all LLM approaches fall behind the tailored connectionist and neuro-symbolic solutions. 
Notably, with only 480 learnable parameters, \name achieves a high accuracy of 98.4\%. 
Moreover, we show that post-programming training allows to extend the knowledge of the model, rather than completely erasing it as shown in other settings~\cite{wu_cognitive_2019}, resulting in a monotonic increase in downstream accuracy.

\subsection{Ablation of LLM prompting techniques}\label{sec:llm-ablation}
Table~\ref{tab:llm-ablation} shows the task accuracy on I-RAVEN using GPT-4 and Llama-3 70B in various prompting configurations. 
Overall, both models benefit from the additional guidance provided by our prompting techniques. 
Concretely, using a predictive approach and querying for individual disentangled attributes yielded already high accuracies (91.4\% and 83.2\% for GPT-4 and Llama-3 70B, respectively). 
Introducing self-consistency further improves the accuracy for both models.
Llama-3 70B's performance can be further pushed (to 85.0\%) by using self-consistency and in-context learning. 
On the contrary, GPT-4 cannot make use of the additional in-context samples, yielding a lower accuracy instead. 

\begin{table*}[t!]
\caption{Ablation study considering various LLM prompting techniques. We report the task accuracy (\%) on the \texttt{center} constellation of I-RAVEN.}
\label{tab:llm-ablation}
\centering
\begin{threeparttable}
\begin{tabular}{lccccc}
\toprule
\begin{tabular}[c]{@{}l@{}}Predictive/\\ discriminative\end{tabular} & \begin{tabular}[c]{@{}l@{}}Disentangled queries \\ per attribute (3$\times$queries)\end{tabular} & \begin{tabular}[c]{@{}l@{}}Self-consistency \\ (n=7)\end{tabular} & \begin{tabular}[c]{@{}l@{}}In-context learning \\ (s=16)\end{tabular} & GPT-4    & Llama-3 70B   \\
\cmidrule(r){1-4}\cmidrule(r){5-6}
Discriminative                                                       &                                                                 &                                                                   &                                                                       & 56.0     & 22.8          \\
Discriminative                                                       & \Included                                                                &                                                                   &                                                                       & 60.0     & 22.4          \\
Predictive                                                           &                                                                    &                                                                   &                                                                       & 74.8     & 79.0          \\
Predictive                                                           & \Included                                                                &                                                                   &                                                                       & 91.4     & 83.2          \\
Predictive                                                           & \Included                                                                & \Included                                                               &                                                                       & \textbf{93.2}     & 84.8          \\
Predictive                                                           & \Included                                                                &                                                                   & \Included                                                                   & 85.4     & 84.8          \\
Predictive                                                           & \Included                                                                & \Included                                                               & \Included                                                                   &  86.4        &     \textbf{85.0}          \\
\bottomrule
\end{tabular}
\end{threeparttable}
\end{table*}

\begin{table*}[tbp]
\caption{Accuracy (\%) of predicting the correct attribute value. Self-consistency (n=7) is used. Results are averaged across all attributes.}
\label{tab:rules}
\centering
\begin{tabular}{lccccc}
\toprule
Model                        & \begin{tabular}[c]{@{}l@{}}Disentangled queries \\ per attribute (3$\times$queries)\end{tabular} & Constant & Progression & Distribute three & Arithmetic \\
\cmidrule(r){1-2}\cmidrule(r){3-6}
\multirow{2}{*}{GPT-4}       & No                                                                                   & 100      & 98.0        & 91.6             & 27.1       \\
                             & Yes                                                                                  & 100      & 100         & 99.5             & 73.6       \\
\cmidrule(r){1-2}\cmidrule(r){3-6}
\multirow{2}{*}{Llama-3 70B} & No                                                                                   & 100      & 97.2        & 99.3             & 31.0       \\
                             & Yes                                                                                  & 100      & 100         & 96.6             & 45.0      \\
\bottomrule
\end{tabular}
\end{table*}

\subsection{LLMs show weakness in arithmetic rule}
Even though both \glspl*{llm} achieve a reasonable overall task accuracy, they fail in some instances. 
We shed more light on the reasoning capability of the two models by analyzing the accuracy of predicting the correct value for a given rule. 
As shown in Table~\ref{tab:rules}, both models perform well on \texttt{constant}, \texttt{progression}, and \texttt{distribute three} rules, whereas the accuracy notably drops for the \texttt{arithmetic} rule. 
One explanation for the accuracy drop could be the \gls*{llm}'s tendency for (short-sighted) relational reasoning, instead of performing relational mapping that requires the understanding of the first two rows before applying a rule on the last row~\cite{stevenson_large_2023}. 
We analyze this hypothesis in Appendix~\ref{app:arithmetic-analysis}, where we attempt to explain the \gls*{llm}'s wrong predictions by rules that may have been inferred from the last row. 
For GPT-4, 32 out of 68 errors can be explained by rules that might have been inferred from a partial context matrix, e.g., a \texttt{constant} or \texttt{progression} rule based on the last row. 

\subsection{Results on our novel I-RAVEN-X}

\begin{table}[t]
    \caption{Task accuracy (\%) on I-RAVEN and our novel I-RAVEN-X. The \glspl*{llm} use self-consistency (n=7). }
    \label{tab:i-raven-x-task}
    \begin{tabular}{lrrrr}
    \toprule
    \multicolumn{1}{l}{} & \multicolumn{1}{c}{I-RAVEN} & \multicolumn{3}{c}{I-RAVEN-X} \\
              & \multicolumn{1}{c}{$3\times3$} & \multicolumn{3}{c}{$3\times10$} \\
    \cmidrule(r){2-2}\cmidrule(r){3-5}
    Dynamic range ($m$)        & \multicolumn{1}{c}{5--10}                       & \multicolumn{1}{c}{50}       & \multicolumn{1}{c}{100}      & \multicolumn{1}{c}{1000}    \\
    \cmidrule(r){1-1} \cmidrule(r){2-2}\cmidrule(r){3-5}
    Llama-3 70B          &   85.0                 & 76.8     & 73.0     & 74.2    \\
    GPT-4                &   93.2                  & 82.2     & 79.6     & 76.6   \\
    $\text{\name}_\text{progr}$ & $97.2$& 100.0 & 100.0         & 99.7\\
    $\text{\name}_\text{learn}$& $98.4$ &  95.0 & 95.0 & 90.6 \\
    \bottomrule
    \end{tabular}
    \end{table}

\begin{table}[t]
\caption{Arithmetic accuracy (\%) on I-RAVEN and our novel I-RAVEN-X. The \glspl*{llm} use self-consistency (n=7).}
\label{tab:i-raven-x-arith}
\begin{tabular}{lrrrr}
\toprule
\multicolumn{1}{l}{} & \multicolumn{1}{c}{I-RAVEN} & \multicolumn{3}{c}{I-RAVEN-X} \\
          & \multicolumn{1}{c}{$3\times3$} & \multicolumn{3}{c}{$3\times10$} \\
\cmidrule(r){2-2}\cmidrule(r){3-5}
Dynamic range ($m$)        & \multicolumn{1}{c}{5--10}                       & \multicolumn{1}{c}{50}       & \multicolumn{1}{c}{100}      & \multicolumn{1}{c}{1000}    \\
\cmidrule(r){1-1} \cmidrule(r){2-2}\cmidrule(r){3-5}
Llama-3 70B          &  45.0                 &  1.5    &  2.6    & 0.4   \\
GPT-4                &  73.6                 &   30.4  &  25.1  & 8.4  \\
$\text{\name}_\text{progr}$   &     100.0         &              99.8               &     100.0     &          99.5       \\
$\text{\name}_\text{learn}$                 &                  99.9          &        81.6  &       78.8   &    61.6    \\
\bottomrule
\end{tabular}
\end{table}

Finally, we conduct experiments on our novel I-RAVEN-X test, which allows us to configure the matrix size and the dynamic range of the attribute values. 
We fix the grid size to $3\times10$ and vary the dynamic range between 50, 100, and 1000. 
As shown in Table~\ref{tab:i-raven-x-task}, the \gls*{llm}'s drops not only due to the larger grid size but also generally degrades with an increasing dynamic range.
At the same time, our \name maintains a high accuracy across the board, while only being trained at dynamic range of 50 and reconfigured for the higher ranges.
Investigating the performance on the \texttt{arithmetic} rule in Table~\ref{tab:i-raven-x-arith} explains the overall accuracy degradation: the arithmetic accuracy drops below 10\% for both \glspl*{llm} at the highest dynamic range (1000). 

\section{Conclusion}
This work revealed \gls*{llm}'s limitations in recognizing and executing arithmetic rules in abstract reasoning tasks, despite being provided disentangled prompts with ground-truth visual attributes and using advanced prompting techniques.
We further showed the serious limitation on a larger (3$\times$10) RPM test. 
As a viable alternative, we presented a neuro-symbolic approach (\name) that achieves a high accuracy both on I-RAVEN and our I-RAVEN-X, thanks to learning to reason with distributed VSA representations and operators. 
We hope that our findings will lead to the development of architectures that aim to improve reasoning capabilities, e.g., by integrating symbolic solvers such as our \name into \glspl*{llm}.

\section*{References}
\bibliography{hd-learning, bib-short}

\begin{thebibliography}{39}
\ifx \bisbn   \undefined \def \bisbn  #1{ISBN #1}\fi
\ifx \binits  \undefined \def \binits#1{#1} \fi
\ifx \bauthor  \undefined \def \bauthor#1{#1} \fi
\ifx \bjtitle  \undefined \def \bjtitle#1{\textit{#1}}\fi
\ifx \batitle  \undefined \def \batitle#1{#1} \fi
\ifx \bctitle  \undefined \def \bctitle#1{#1} \fi
\ifx \bvolume  \undefined \def \bvolume#1{\textbf{#1}}\fi
\ifx \byear  \undefined \def \byear#1{#1} \fi
\ifx \bissue  \undefined \def \bissue#1{#1} \fi
\ifx \bfpage  \undefined \def \bfpage#1{#1} \fi
\ifx \blpage  \undefined \def \blpage #1{#1} \fi
\ifx \burl  \undefined \def \burl#1{#1} \fi
\ifx \doiurl  \undefined \def \doiurl#1{#1} \fi
\ifx \betal  \undefined \def \betal{et al.} \fi
\ifx \binstitute  \undefined \def \binstitute#1{#1} \fi
\ifx \beditor  \undefined \def \beditor#1{#1} \fi
\ifx \bpublisher  \undefined \def \bpublisher#1{#1} \fi
\ifx \bbtitle  \undefined \def \bbtitle#1{\textit{#1}} \fi
\ifx \bedition  \undefined \def \bedition#1{#1} \fi
\ifx \bseriesno  \undefined \def \bseriesno#1{#1} \fi
\ifx \blocation  \undefined \def \blocation#1{#1} \fi
\ifx \bsertitle  \undefined \def \bsertitle#1{#1} \fi
\ifx \bsnm \undefined \def \bsnm#1{#1} \fi
\ifx \bsuffix \undefined \def \bsuffix#1{#1} \fi
\ifx \bparticle \undefined \def \bparticle#1{#1} \fi
\ifx \barticle \undefined \def \barticle#1{#1} \fi
\ifx \botherref \undefined \def \botherref #1{#1} \fi
\ifx \url \undefined \def \url#1{#1} \fi
\ifx \bchapter \undefined \def \bchapter#1{#1} \fi
\ifx \bbook \undefined \def \bbook#1{#1} \fi
\ifx \bcomment \undefined \def \bcomment#1{#1} \fi
\ifx \oauthor \undefined \def \oauthor#1{#1} \fi
\ifx \citeauthoryear \undefined \def \citeauthoryear#1{#1} \fi
\ifx \texttildelow  \undefined \def \texttildelow{\symbol{126}} \fi
\def \endbibitem {}
\ifx \bconflocation  \undefined \def \bconflocation#1{#1} \fi

\bibitem{barrett_measuring_2018}
\begin{bchapter}
\bauthor{\binits{D.G.T.}~\bsnm{Barrett}},
\bauthor{\binits{F.}~\bsnm{Hill}},
\bauthor{\binits{A.}~\bsnm{Santoro}},
\bauthor{\binits{A.S.}~\bsnm{Morcos}} and
\bauthor{\binits{T.}~\bsnm{Lillicrap}},
\bctitle{Measuring abstract reasoning in neural networks},
in: \bbtitle{International {Conference} on {Machine} {Learning} ({ICML})},
\byear{2018},
pp.~\bfpage{511}--\blpage{520}.
\end{bchapter}
\endbibitem

\bibitem{zhang_raven_2019}
\begin{bchapter}
\bauthor{\binits{C.}~\bsnm{Zhang}},
\bauthor{\binits{F.}~\bsnm{Gao}},
\bauthor{\binits{B.}~\bsnm{Jia}},
\bauthor{\binits{Y.}~\bsnm{Zhu}} and
\bauthor{\binits{S.-C.}~\bsnm{Zhu}},
\bctitle{{RAVEN}: {A} {Dataset} for {Relational} and {Analogical} {Visual} {REasoNing}},
in: \bbtitle{2019 {IEEE}/{CVF} {Conference} on {Computer} {Vision} and {Pattern} {Recognition} ({CVPR})},
\bpublisher{IEEE},
\blocation{Long Beach, CA, USA},
\byear{2019},
pp.~\bfpage{5312}--\blpage{5322}.
\end{bchapter}
\endbibitem

\bibitem{hu_stratified_2021}
\begin{barticle}
\bauthor{\binits{S.}~\bsnm{Hu}},
\bauthor{\binits{Y.}~\bsnm{Ma}},
\bauthor{\binits{X.}~\bsnm{Liu}},
\bauthor{\binits{Y.}~\bsnm{Wei}} and
\bauthor{\binits{S.}~\bsnm{Bai}},
\batitle{Stratified {Rule}-{Aware} {Network} for {Abstract} {Visual} {Reasoning}},
\bjtitle{Proceedings of the AAAI Conference on Artificial Intelligence}
\bvolume{35}(\bissue{2})
(\byear{2021}),
\bfpage{1567}--\blpage{1574}.
\end{barticle}
\endbibitem

\bibitem{benny_scale-localized_2021}
\begin{bchapter}
\bauthor{\binits{Y.}~\bsnm{Benny}},
\bauthor{\binits{N.}~\bsnm{Pekar}} and
\bauthor{\binits{L.}~\bsnm{Wolf}},
\bctitle{Scale-{Localized} {Abstract} {Reasoning}},
in: \bbtitle{2021 {IEEE}/{CVF} {Conference} on {Computer} {Vision} and {Pattern} {Recognition} ({CVPR})},
\bpublisher{IEEE},
\blocation{Nashville, TN, USA},
\byear{2021},
pp.~\bfpage{12552}--\blpage{12560}.
\end{bchapter}
\endbibitem

\bibitem{wu_scattering_2020}
\begin{botherref}
\oauthor{\binits{Y.}~\bsnm{Wu}},
\oauthor{\binits{H.}~\bsnm{Dong}},
\oauthor{\binits{R.}~\bsnm{Grosse}} and
\oauthor{\binits{J.}~\bsnm{Ba}},
The {Scattering} {Compositional} {Learner}: {Discovering} {Objects}, {Attributes}, {Relationships} in {Analogical} {Reasoning},
\textit{arxiv preprint arXiv:2007.04212}
(2020).
\end{botherref}
\endbibitem

\bibitem{zhang_abstract_2021}
\begin{bchapter}
\bauthor{\binits{C.}~\bsnm{Zhang}},
\bauthor{\binits{B.}~\bsnm{Jia}},
\bauthor{\binits{S.-C.}~\bsnm{Zhu}} and
\bauthor{\binits{Y.}~\bsnm{Zhu}},
\bctitle{Abstract {Spatial}-{Temporal} {Reasoning} via {Probabilistic} {Abduction} and {Execution}},
in: \bbtitle{2021 {IEEE}/{CVF} {Conference} on {Computer} {Vision} and {Pattern} {Recognition} ({CVPR})},
\bpublisher{IEEE},
\blocation{Nashville, TN, USA},
\byear{2021},
pp.~\bfpage{9731}--\blpage{9741}.
\end{bchapter}
\endbibitem

\bibitem{hersche_neuro-vector-symbolic_2023}
\begin{barticle}
\bauthor{\binits{M.}~\bsnm{Hersche}},
\bauthor{\binits{M.}~\bsnm{Zeqiri}},
\bauthor{\binits{L.}~\bsnm{Benini}},
\bauthor{\binits{A.}~\bsnm{Sebastian}} and
\bauthor{\binits{A.}~\bsnm{Rahimi}},
\batitle{A {Neuro}-vector-symbolic {Architecture} for {Solving} {Raven}'s {Progressive} {Matrices}},
\bjtitle{Nature Machine Intelligence}
\bvolume{5}(\bissue{4})
(\byear{2023}),
\bfpage{363}--\blpage{375}.
\end{barticle}
\endbibitem

\bibitem{hersche_probabilistic_2024}
\begin{bchapter}
\bauthor{\binits{M.}~\bsnm{Hersche}},
\bauthor{\binits{F.}~\bsnm{di~Stefano}},
\bauthor{\binits{T.}~\bsnm{Hofmann}},
\bauthor{\binits{A.}~\bsnm{Sebastian}} and
\bauthor{\binits{A.}~\bsnm{Rahimi}},
\bctitle{Probabilistic {Abduction} for {Visual} {Abstract} {Reasoning} via {Learning} {Rules} in {Vector}-symbolic {Architectures}},
in: \bbtitle{The 3rd {Workshop} on {Mathematical} {Reasoning} and {AI} at {NeurIPS}'23},
\byear{2024}.
\end{bchapter}
\endbibitem

\bibitem{camposampiero_towards_2024}
\begin{bchapter}
\bauthor{\binits{G.}~\bsnm{Camposampiero}},
\bauthor{\binits{M.}~\bsnm{Hersche}},
\bauthor{\binits{A.}~\bsnm{Terzic}},
\bauthor{\binits{R.}~\bsnm{Wattenhofer}},
\bauthor{\binits{A.}~\bsnm{Sebastian}} and
\bauthor{\binits{A.}~\bsnm{Rahimi}},
\bctitle{Towards {Learning} {Abductive} {Reasoning} using {VSA} {Distributed} {Representations}},
in: \bbtitle{International {Conference} on {Neural}-{Symbolic} {Learning} and {Reasoning}},
\bpublisher{Springer},
\byear{2024},
pp.~\bfpage{370}--\blpage{385}.
\end{bchapter}
\endbibitem

\bibitem{webb_emergent_2023}
\begin{barticle}
\bauthor{\binits{T.}~\bsnm{Webb}},
\bauthor{\binits{K.J.}~\bsnm{Holyoak}} and
\bauthor{\binits{H.}~\bsnm{Lu}},
\batitle{Emergent analogical reasoning in large language models},
\bjtitle{Nature Human Behaviour}
\bvolume{7}(\bissue{9})
(\byear{2023}),
\bfpage{1526}--\blpage{1541}.
\end{barticle}
\endbibitem

\bibitem{stevenson_large_2023}
\begin{botherref}
\oauthor{\binits{C.E.}~\bsnm{Stevenson}},
\oauthor{\binits{M.}~\bsnm{ter Veen}},
\oauthor{\binits{R.}~\bsnm{Choenni}},
\oauthor{\binits{H.L.J.}~\bsnm{van~der Maas}} and
\oauthor{\binits{E.}~\bsnm{Shutova}},
Do large language models solve verbal analogies like children do?,
\textit{arxiv preprint arXiv:2310.20384}
(2023).
\end{botherref}
\endbibitem

\bibitem{gendron_large_2024-1}
\begin{bchapter}
\bauthor{\binits{G.}~\bsnm{Gendron}},
\bauthor{\binits{Q.}~\bsnm{Bao}},
\bauthor{\binits{M.}~\bsnm{Witbrock}} and
\bauthor{\binits{G.}~\bsnm{Dobbie}},
\bctitle{Large {Language} {Models} {Are} {Not} {Strong} {Abstract} {Reasoners}},
in: \bbtitle{Thirty-{Third} {International} {Joint} {Conference} on {Artificial} {Intelligence} ({IJCAI})},
Vol.~\bseriesno{7},
\byear{2024},
pp.~\bfpage{6270}--\blpage{6278},
\bcomment{ISSN: 1045-0823}.
\end{bchapter}
\endbibitem

\bibitem{lewis_evaluating_2024}
\begin{barticle}
\bauthor{\binits{M.}~\bsnm{Lewis}} and
\bauthor{\binits{M.}~\bsnm{Mitchell}},
\batitle{Evaluating the {Robustness} of {Analogical} {Reasoning} in {Large} {Language} {Models}},
\bjtitle{arXiv preprint arXiv:2411.14215}
(\byear{2024}).
\end{barticle}
\endbibitem

\bibitem{cao_what_2024}
\begin{botherref}
\oauthor{\binits{X.}~\bsnm{Cao}},
\oauthor{\binits{B.}~\bsnm{Lai}},
\oauthor{\binits{W.}~\bsnm{Ye}},
\oauthor{\binits{Y.}~\bsnm{Ma}},
\oauthor{\binits{J.}~\bsnm{Heintz}},
\oauthor{\binits{J.}~\bsnm{Chen}},
\oauthor{\binits{J.}~\bsnm{Cao}} and
\oauthor{\binits{J.M.}~\bsnm{Rehg}},
What is the {Visual} {Cognition} {Gap} between {Humans} and {Multimodal} {LLMs}?,
\textit{arXiv preprint arXiv:2406.10424}
(2024).
\end{botherref}
\endbibitem

\bibitem{hu_-context_2023}
\begin{bchapter}
\bauthor{\binits{X.}~\bsnm{Hu}},
\bauthor{\binits{S.}~\bsnm{Storks}},
\bauthor{\binits{R.}~\bsnm{Lewis}} and
\bauthor{\binits{J.}~\bsnm{Chai}},
\bctitle{In-{Context} {Analogical} {Reasoning} with {Pre}-{Trained} {Language} {Models}},
in: \bbtitle{Proceedings of the 61st {Annual} {Meeting} of the {Association} for {Computational} {Linguistics} ({Volume} 1: {Long} {Papers})},
\bpublisher{Association for Computational Linguistics},
\blocation{Toronto, Canada},
\byear{2023},
pp.~\bfpage{1953}--\blpage{1969}.
\end{bchapter}
\endbibitem

\bibitem{mitchell_comparing_2024}
\begin{bchapter}
\bauthor{\binits{M.}~\bsnm{Mitchell}},
\bauthor{\binits{A.B.}~\bsnm{Palmarini}} and
\bauthor{\binits{A.}~\bsnm{Moskvichev}},
\bctitle{Comparing {Humans}, {GPT}-4, and {GPT}-{4V} {On} {Abstraction} and {Reasoning} {Tasks}},
in: \bbtitle{{AAAI} 2024 {Workshop} on ''{Are} {Large} {Language} {Models} {Simply} {Causal} {Parrots}?''},
\byear{2024}.
\end{bchapter}
\endbibitem

\bibitem{camposampiero_abstract_2023}
\begin{bchapter}
\bauthor{\binits{G.}~\bsnm{Camposampiero}},
\bauthor{\binits{L.}~\bsnm{Houmard}},
\bauthor{\binits{B.}~\bsnm{Estermann}},
\bauthor{\binits{J.}~\bsnm{Mathys}} and
\bauthor{\binits{R.}~\bsnm{Wattenhofer}},
\bctitle{Abstract {Visual} {Reasoning} {Enabled} by {Language}},
in: \bbtitle{2023 {IEEE}/{CVF} {Conference} on {Computer} {Vision} and {Pattern} {Recognition} {Workshops} ({CVPRW})},
\bpublisher{IEEE},
\blocation{Vancouver, BC, Canada},
\byear{2023},
pp.~\bfpage{2643}--\blpage{2647}.
\end{bchapter}
\endbibitem

\bibitem{jiang_marvel_2024}
\begin{botherref}
\oauthor{\binits{Y.}~\bsnm{Jiang}},
\oauthor{\binits{J.}~\bsnm{Zhang}},
\oauthor{\binits{K.}~\bsnm{Sun}},
\oauthor{\binits{Z.}~\bsnm{Sourati}},
\oauthor{\binits{K.}~\bsnm{Ahrabian}},
\oauthor{\binits{K.}~\bsnm{Ma}},
\oauthor{\binits{F.}~\bsnm{Ilievski}} and
\oauthor{\binits{J.}~\bsnm{Pujara}},
{MARVEL}: {Multidimensional} {Abstraction} and {Reasoning} through {Visual} {Evaluation} and {Learning},
\textit{arXiv preprint arXiv:2404.13591}
(2024).
\end{botherref}
\endbibitem

\bibitem{ahrabian_curious_2024}
\begin{barticle}
\bauthor{\binits{K.}~\bsnm{Ahrabian}},
\bauthor{\binits{Z.}~\bsnm{Sourati}},
\bauthor{\binits{K.}~\bsnm{Sun}},
\bauthor{\binits{J.}~\bsnm{Zhang}},
\bauthor{\binits{Y.}~\bsnm{Jiang}},
\bauthor{\binits{F.}~\bsnm{Morstatter}} and
\bauthor{\binits{J.}~\bsnm{Pujara}},
\batitle{The {Curious} {Case} of {Nonverbal} {Abstract} {Reasoning} with {Multi}-{Modal} {Large} {Language} {Models}},
\bjtitle{arXiv preprint arXiv:2401.12117}
(\byear{2024}).
\end{barticle}
\endbibitem

\bibitem{zhang_how_2024}
\begin{bchapter}
\bauthor{\binits{Y.}~\bsnm{Zhang}},
\bauthor{\binits{H.}~\bsnm{Bai}},
\bauthor{\binits{R.}~\bsnm{Zhang}},
\bauthor{\binits{J.}~\bsnm{Gu}},
\bauthor{\binits{S.}~\bsnm{Zhai}},
\bauthor{\binits{J.M.}~\bsnm{Susskind}} and
\bauthor{\binits{N.}~\bsnm{Jaitly}},
\bctitle{How {Far} {Are} {We} from {Intelligent} {Visual} {Deductive} {Reasoning}?},
in: \bbtitle{{ICLR} 2024 {Workshop}: {How} {Far} {Are} {We} {From} {AGI}},
\byear{2024}.
\end{bchapter}
\endbibitem

\bibitem{wust_bongard_2024}
\begin{bchapter}
\bauthor{\binits{A.}~\bsnm{Wüst}},
\bauthor{\binits{T.}~\bsnm{Tobiasch}},
\bauthor{\binits{L.}~\bsnm{Helff}},
\bauthor{\binits{D.S.}~\bsnm{Dhami}},
\bauthor{\binits{C.A.}~\bsnm{Rothkopf}} and
\bauthor{\binits{K.}~\bsnm{Kersting}},
\bctitle{Bongard in {Wonderland}: {Visual} {Puzzles} that {Still} {Make} {AI} {Go} {Mad}?},
in: \bbtitle{The {First} {Workshop} on {System}-2 {Reasoning} at {Scale}, {NeurIPS}'24},
\byear{2024}.
\end{bchapter}
\endbibitem

\bibitem{latif_systematic_2024_short}
\begin{barticle}
\bauthor{\binits{E.}~\bsnm{Latif}},
\bauthor{\bsnm{{Yifan Zhou}}},
\bauthor{\bsnm{{Shuchen Guo}}},
\bauthor{\bsnm{{Yizhu Gao}}},
\bauthor{\bsnm{{Lehong Shi}}},
\bauthor{\binits{M.}~\bsnm{Nyaaba}},
\bauthor{\bsnm{{Gyeonggeon Lee}}},
\bauthor{\binits{L.}~\bsnm{Zhang}} \betal,
\batitle{A {Systematic} {Assessment} of {OpenAI} o1-{Preview} for {Higher} {Order} {Thinking} in {Education}},
\bjtitle{arXiv preprint arXiv:2410.21287}
(\byear{2024}),
\bcomment{Publisher: Unpublished}.
\end{barticle}
\endbibitem

\bibitem{raven_ravens_1938}
\begin{bbook}
\bauthor{\binits{J.C.}~\bsnm{Raven}},
\bauthor{\binits{J.H.}~\bsnm{Court}} and
\bauthor{\binits{J.}~\bsnm{Raven}},
\bbtitle{Raven's progressive matrices},
\bpublisher{Oxford Psychologists Press},
\byear{1938}.
\end{bbook}
\endbibitem

\bibitem{plate_holographic_1995}
\begin{botherref}
\oauthor{\binits{T.A.}~\bsnm{Plate}},
Holographic {Reduced} {Representations},
\textit{IEEE Transactions on Neural Networks and Learning Systems}
\textbf{6}(3)
(1995).
\end{botherref}
\endbibitem

\bibitem{plate_holographic_2003}
\begin{bbook}
\bauthor{\binits{T.A.}~\bsnm{Plate}},
\bbtitle{Holographic {Reduced} {Representations}: {Distributed} {Representation} for {Cognitive} {Structures}},
\bpublisher{Center for the Study of Language and Information, Stanford},
\byear{2003}.
\end{bbook}
\endbibitem

\bibitem{gayler_vector_2003}
\begin{bchapter}
\bauthor{\binits{R.W.}~\bsnm{Gayler}},
\bctitle{Vector {Symbolic} {Architectures} answer {Jackendoff}'s challenges for cognitive neuroscience},
in: \bbtitle{Proceedings of the {Joint} {International} {Conference} on {Cognitive} {Science}. {ICCS}/{ASCS}},
\byear{2003},
pp.~\bfpage{133}--\blpage{138}.
\end{bchapter}
\endbibitem

\bibitem{kanerva_hyperdimensional_2009}
\begin{barticle}
\bauthor{\binits{P.}~\bsnm{Kanerva}},
\batitle{Hyperdimensional {Computing}: {An} {Introduction} to {Computing} in {Distributed} {Representation} with {High}-{Dimensional} {Random} {Vectors}},
\bjtitle{Cognitive Computation}
\bvolume{1}(\bissue{2})
(\byear{2009}),
\bfpage{139}--\blpage{159}.
\end{barticle}
\endbibitem

\bibitem{openai_gpt-4_2024_short}
\begin{botherref}
\oauthor{\bsnm{OpenAI}},
\oauthor{\binits{J.}~\bsnm{Achiam}},
\oauthor{\binits{S.}~\bsnm{Adler}},
\oauthor{\binits{S.}~\bsnm{Agarwal}},
\oauthor{\binits{L.}~\bsnm{Ahmad}},
\oauthor{\binits{I.}~\bsnm{Akkaya}},
\oauthor{\binits{F.L.}~\bsnm{Aleman}},
\oauthor{\binits{D.}~\bsnm{Almeida}},
\oauthor{\binits{J.}~\bsnm{Altenschmidt}},
\oauthor{\binits{S.}~\bsnm{Altman}} \betal,
{GPT}-4 {Technical} {Report},
\textit{arXiv preprint arXiv:2303.08774}
(2024).
\end{botherref}
\endbibitem

\bibitem{dubey_llama_2024_short}
\begin{barticle}
\bauthor{\binits{A.}~\bsnm{Dubey}},
\bauthor{\binits{A.}~\bsnm{Jauhri}},
\bauthor{\binits{A.}~\bsnm{Pandey}},
\bauthor{\binits{A.}~\bsnm{Kadian}},
\bauthor{\binits{A.}~\bsnm{Al-Dahle}},
\bauthor{\binits{A.}~\bsnm{Letman}},
\bauthor{\binits{A.}~\bsnm{Mathur}},
\bauthor{\binits{A.}~\bsnm{Schelten}} \betal,
\batitle{The {Llama} 3 {Herd} of {Models}},
\bjtitle{arxiv preprint arXiv:2407.21783}
(\byear{2024}).
\end{barticle}
\endbibitem

\bibitem{holyoak_oxford_2013}
\begin{bbook}
\bauthor{\binits{K.J.}~\bsnm{Holyoak}} and
\bauthor{\binits{R.G.}~\bsnm{Morrison}},
\bbtitle{The {Oxford} {Handbook} of {Thinking} and {Reasoning}},
\bpublisher{OUP USA},
\byear{2013}.
\end{bbook}
\endbibitem

\bibitem{wang_self-consistency_2023}
\begin{bchapter}
\bauthor{\binits{X.}~\bsnm{Wang}},
\bauthor{\binits{J.}~\bsnm{Wei}},
\bauthor{\binits{D.}~\bsnm{Schuurmans}},
\bauthor{\binits{Q.}~\bsnm{Le}},
\bauthor{\binits{E.H.}~\bsnm{Chi}},
\bauthor{\binits{S.}~\bsnm{Narang}},
\bauthor{\binits{A.}~\bsnm{Chowdhery}} and
\bauthor{\binits{D.}~\bsnm{Zhou}},
\bctitle{Self-{Consistency} {Improves} {Chain} of {Thought} {Reasoning} in {Language} {Models}},
in: \bbtitle{The {Eleventh} {International} {Conference} on {Learning} {Representations} ({ICLR})},
\byear{2023}.
\end{bchapter}
\endbibitem

\bibitem{lewkowycz_solving_2022}
\begin{bchapter}
\bauthor{\binits{A.}~\bsnm{Lewkowycz}},
\bauthor{\binits{A.}~\bsnm{Andreassen}},
\bauthor{\binits{D.}~\bsnm{Dohan}},
\bauthor{\binits{E.}~\bsnm{Dyer}},
\bauthor{\binits{H.}~\bsnm{Michalewski}},
\bauthor{\binits{V.}~\bsnm{Ramasesh}},
\bauthor{\binits{A.}~\bsnm{Slone}},
\bauthor{\binits{C.}~\bsnm{Anil}},
\bauthor{\binits{I.}~\bsnm{Schlag}},
\bauthor{\binits{T.}~\bsnm{Gutman-Solo}},
\bauthor{\binits{Y.}~\bsnm{Wu}},
\bauthor{\binits{B.}~\bsnm{Neyshabur}},
\bauthor{\binits{G.}~\bsnm{Gur-Ari}} and
\bauthor{\binits{V.}~\bsnm{Misra}},
\bctitle{Solving {Quantitative} {Reasoning} {Problems} {With} {Language} {Models}},
in: \bbtitle{Advances in {Neural} {Information} {Processing} {Systems} ({NeurIPS})},
Vol.~\bseriesno{35},
\byear{2022},
pp.~\bfpage{3843}--\blpage{3857}.
\end{bchapter}
\endbibitem

\bibitem{brown_language_2020}
\begin{botherref}
\oauthor{\binits{T.B.}~\bsnm{Brown}},
\oauthor{\binits{B.}~\bsnm{Mann}},
\oauthor{\binits{N.}~\bsnm{Ryder}},
\oauthor{\binits{M.}~\bsnm{Subbiah}},
\oauthor{\binits{J.}~\bsnm{Kaplan}},
\oauthor{\binits{P.}~\bsnm{Dhariwal}},
\oauthor{\binits{A.}~\bsnm{Neelakantan}},
\oauthor{\binits{P.}~\bsnm{Shyam}},
\oauthor{\binits{G.}~\bsnm{Sastry}},
\oauthor{\binits{A.}~\bsnm{Askell}},
\oauthor{\binits{S.}~\bsnm{Agarwal}},
\oauthor{\binits{A.}~\bsnm{Herbert-Voss}},
\oauthor{\binits{G.}~\bsnm{Krueger}},
\oauthor{\binits{T.}~\bsnm{Henighan}},
\oauthor{\binits{R.}~\bsnm{Child}},
\oauthor{\binits{A.}~\bsnm{Ramesh}},
\oauthor{\binits{D.M.}~\bsnm{Ziegler}},
\oauthor{\binits{J.}~\bsnm{Wu}},
\oauthor{\binits{C.}~\bsnm{Winter}},
\oauthor{\binits{C.}~\bsnm{Hesse}},
\oauthor{\binits{M.}~\bsnm{Chen}},
\oauthor{\binits{E.}~\bsnm{Sigler}},
\oauthor{\binits{M.}~\bsnm{Litwin}},
\oauthor{\binits{S.}~\bsnm{Gray}},
\oauthor{\binits{B.}~\bsnm{Chess}},
\oauthor{\binits{J.}~\bsnm{Clark}},
\oauthor{\binits{C.}~\bsnm{Berner}},
\oauthor{\binits{S.}~\bsnm{McCandlish}},
\oauthor{\binits{A.}~\bsnm{Radford}},
\oauthor{\binits{I.}~\bsnm{Sutskever}} and
\oauthor{\binits{D.}~\bsnm{Amodei}},
Language {Models} are {Few}-{Shot} {Learners},
\textit{arxiv preprint arXiv:2005.14165}
(2020).
\end{botherref}
\endbibitem

\bibitem{frady_computing_2022}
\begin{bchapter}
\bauthor{\binits{E.P.}~\bsnm{Frady}},
\bauthor{\binits{D.}~\bsnm{Kleyko}},
\bauthor{\binits{C.J.}~\bsnm{Kymn}},
\bauthor{\binits{B.A.}~\bsnm{Olshausen}} and
\bauthor{\binits{F.T.}~\bsnm{Sommer}},
\bctitle{Computing on {Functions} {Using} {Randomized} {Vector} {Representations} (in brief)},
in: \bbtitle{Neuro-{Inspired} {Computational} {Elements} {Conference}},
\bpublisher{ACM},
\blocation{Virtual Event USA},
\byear{2022},
pp.~\bfpage{115}--\blpage{122}.
\end{bchapter}
\endbibitem

\bibitem{hersche_factorizers_2024}
\begin{botherref}
\oauthor{\binits{M.}~\bsnm{Hersche}},
\oauthor{\binits{A.}~\bsnm{Terzic}},
\oauthor{\binits{G.}~\bsnm{Karunaratne}},
\oauthor{\binits{J.}~\bsnm{Langenegger}},
\oauthor{\binits{A.}~\bsnm{Pouget}},
\oauthor{\binits{G.}~\bsnm{Cherubini}},
\oauthor{\binits{L.}~\bsnm{Benini}},
\oauthor{\binits{A.}~\bsnm{Sebastian}} and
\oauthor{\binits{A.}~\bsnm{Rahimi}},
Factorizers for {Distributed} {Sparse} {Block} {Codes},
\textit{Neurosymbolic Artificial Intelligence}
(2024).
\end{botherref}
\endbibitem

\bibitem{chalmers_high-level_1992}
\begin{barticle}
\bauthor{\binits{D.J.}~\bsnm{Chalmers}},
\bauthor{\binits{R.M.}~\bsnm{French}} and
\bauthor{\binits{D.R.}~\bsnm{Hofstadter}},
\batitle{High-level perception, representation, and analogy: {A} critique of artificial intelligence methodology},
\bjtitle{Journal of Experimental \& Theoretical Artificial Intelligence}
\bvolume{4}(\bissue{3})
(\byear{1992}),
\bfpage{185}--\blpage{211}.
\end{barticle}
\endbibitem

\bibitem{cheng_context-dependent_1990}
\begin{bchapter}
\bauthor{\binits{Y.}~\bsnm{Cheng}},
\bctitle{Context-dependent similarity},
in: \bbtitle{Proceedings of the {Sixth} {Annual} {Conference} on {Uncertainty} in {Artificial} {Intelligence}},
\bpublisher{Elsevier Science Inc.},
\byear{1990},
pp.~\bfpage{41}--\blpage{50}.
\end{bchapter}
\endbibitem

\bibitem{wu_cognitive_2019}
\begin{barticle}
\bauthor{\binits{X.}~\bsnm{Wu}},
\bauthor{\binits{X.}~\bsnm{Zhang}} and
\bauthor{\binits{X.}~\bsnm{Shu}},
\batitle{Cognitive {Deficit} of {Deep} {Learning} in {Numerosity}},
\bjtitle{Proceedings of the AAAI Conference on Artificial Intelligence}
\bvolume{33}(\bissue{01})
(\byear{2019}),
\bfpage{1303}--\blpage{1310}.
\end{barticle}
\endbibitem

\bibitem{kleyko_survey_2023}
\begin{barticle}
\bauthor{\binits{D.}~\bsnm{Kleyko}},
\bauthor{\binits{D.A.}~\bsnm{Rachkovskij}},
\bauthor{\binits{E.}~\bsnm{Osipov}} and
\bauthor{\binits{A.}~\bsnm{Rahimi}},
\batitle{A {Survey} on {Hyperdimensional} {Computing} aka {Vector} {Symbolic} {Architectures}, {Part} {I}: {Models} and {Data} {Transformations}},
\bjtitle{ACM Computing Surveys}
\bvolume{55}(\bissue{6})
(\byear{2023}),
\bfpage{1}--\blpage{40}.
\end{barticle}
\endbibitem

\end{thebibliography}
\bibliographystyle{ios1} 

\appendix

\newpage
\renewcommand\thefigure{\thesection.\arabic{figure}}
\renewcommand\thetable{\thesection.\arabic{table}}   

\section{Prompting details}\label{app:prompting}

This appendix provides more details on our prompting strategy. 
While the prompt design was mainly inspired by~\cite{hu_-context_2023}, we extended it with predictive and discriminative classification and fine-tuned it for the different models.
For example, we found that adding a prefix (``Only return the missing number'') helped to slightly improve GPT4's accuracy, whereas it reduced Llama-3 70B's performance. 
Thus, we used individual prompts for the different models. 

\subsection{Joint attribute querying}
\label{app:disentanglement}
\begin{figure*}[!ht]
    \centering
    \includegraphics[width=0.49\textwidth]{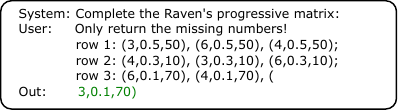}
    \caption{Example prompt for joint prediction of all three attributes.}
    \label{fig-app:joint-prompt}
\end{figure*}

As an alternative to individually querying the \gls*{llm} for predicting the separate attributes, we also devised a joint attribute prompting scheme, shown in \fig~\ref{fig-app:joint-prompt}. 
The attributes of each panel are represented in brackets: (\texttt{shape}, \texttt{size}, \texttt{color}). 
In this setting, the \gls*{llm} is required to predict all three attributes of the missing panel at once.
For better distinguishing between the different attributes, they are scaled with individual factors (1$\times$, 0.1$\times$, 10$\times$). 

\subsection{Discriminative classification approach}
\label{app:discriminative}
\begin{figure*}[!ht]
    \centering
    \includegraphics[width=0.5\textwidth]{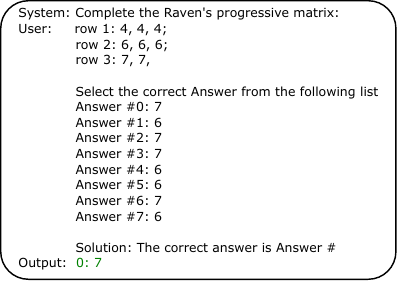}
    \caption{Example prompt for discriminative classification approach, where the answer candidates are provided. The underlying attribute is \texttt{shape} and the rule is \texttt{constant}.}
    \label{fig-app:discriminative-prompts}
\end{figure*}

\fig~\ref{fig-app:discriminative-prompts} shows an example prompt for performing discriminative classification. 
As shown, the answers only contain two distinct values (``6'' and ``7''); finding the correct answer requires the consideration of all attributes.
For choosing the final answer, we extract all attribute values that correspond to the predicted answer (e.g., value ``7'' for \texttt{shape}) and select the best matching answer candidate, i.e., the answer with the highest number of overlaps with the predicted attributes.

\section{Vector-symbolic architectures}\label{app:vsa}

Vector-symbolic architectures (VSAs)~\cite{plate_holographic_1995, plate_holographic_2003,gayler_vector_2003, kanerva_hyperdimensional_2009} are a family of computational models that rely on the mathematical properties of high-dimensional vector spaces.
VSAs make use of high-dimensional distributed representations for structured (symbolic) representation of data while maintaining the advantages of connectionist distributed vector representations (see~\cite{kleyko_survey_2023} for a survey).
Here is a formal definition of VSAs:

\begin{definition}[VSA]
\label{def:vsa}
A vector-symbolic architecture (VSA) consists of a 4-tuple $\mathbb{V}=(\mathbb{C}, \oplus, \otimes,\odot)$, where $\mathbb{C}$ is a set of high-dimensional distributed vectors equipped with two main operations, $\oplus$ (bundling) and $\otimes$ (binding), and on which it is possible to define a similarity measure $\odot$.
\end{definition}
Bundling is a similarity-preserving operation that creates a superposition of the operands, that is, the resulting vector will have a high similarity with the two operands.
Binding, on the other hand, is an operation that allows to bind a vector (value) to another vector (key) and does not preserve similarities; it usually allows an inverse operation, called unbinding.
The specific realization of the bundling, binding, and vector space constitute the main difference between members of the VSA family.

\section{Analysis of arithmetic errors}\label{app:arithmetic-analysis}
This appendix aims to find explanations for \gls*{llm}'s errors by analyzing the structure behind the predicted answers. 
A recent study~\cite{stevenson_large_2023} showed that \glspl*{llm} tend to solve verbal analogy problems in an associative way instead of performing proper relational mapping. 
The associative reasoning can be explained as ignoring the source domain and solving the task directly at the target domain (e.g., only looking at the possible solutions without reading the questions). 
Interestingly, children tend to perform associative reasoning, whereas adults opt for relational mapping. 

In \glspl*{rpm}, the source domain can be defined as the first two rows (with values $x_{1,1}, x_{1,2}, x_{1,3}$ and $x_{2,1}, x_{2,2}, x_{2,3}$), whereby the target domain is the last row ($x_{3,1}, x_{3,2}$). 
Therefore, an associative reasoner would only look at the last row to solve the task. 
In the following, we aim to find potential incorrect rules that the \glspl*{llm} may have been inferred from the last row(s): 
\begin{itemize}
    \item \texttt{constant}: The values of the last row are identical ($x_{3,1}=x_{3,2}$), and the model predicts $\hat{x}_{3,3} =x_{3,2}=x_{3,1}$
    \item \texttt{progression}: The values of the last row differ by $\delta$ = $x_{3,2}-x_{3,1}$, and the model predicts $\hat{x}_{3,3} =x_{3,2} + \delta$
    \item \texttt{short constant}: The model just copies the penultimate value: $\hat{x}_{3,3} =x_{3,2}$. 
    \item \texttt{short distribute three}: Assuming a distribute three over the last two rows: ${x}_{3,1} \in \lbrace x_{2,1}, x_{2,2}, x_{2,3} \rbrace $, ${x}_{3,2} \in \lbrace x_{2,1}, x_{2,2}, x_{2,3} \rbrace $, and hence $\hat{x}_{3,3} \in \lbrace x_{2,1}, x_{2,2}, x_{2,3} \rbrace $. 
\end{itemize}

\begin{figure*}[!ht]
    \centering
    \includegraphics[width=\textwidth]{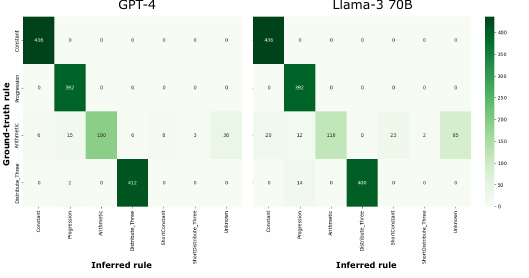}
    \caption{Rule confusion matrix of GPT-4 (left) and Llama-3 70B (right).}
    \label{fig-app:rules}
\end{figure*}

\fig~\ref{fig-app:rules} shows the resulting confusion matrix summarizing all the attributes. 
The \texttt{arithmetic} rule has fewer occurrences as this rule is not integrated in the attribute \texttt{shape}. 
As already stated in the main text, the majority of wrong predictions are related to the \texttt{arithmetic} rule. 
For GPT-4, our new rule interpretations can explain 32 out of the 68 errors, while 36 errors remain unknown. 
Llama-3 70B showed many more errors in the arithmetic rule; here, we can explain 57 out of 142 errors with relational reasoning. 
In summary, some (40.1--47.1\%) of the LLM's errors can be rooted in relational reasoning. 
Further understanding the behavior of the unknown rules is scope for future work. 

\end{document}